\documentclass[runningheads]{llncs}
\usepackage{graphicx}
\usepackage{tikz}
\usepackage{comment}
\usepackage{amsmath,amssymb} % define this before the line numbering.
\usepackage{color}
\usepackage{booktabs}
\usepackage{caption}
\usepackage{subcaption}
\usepackage{multirow}
\usepackage{float}
\definecolor{hotpink}{RGB}{0,180,0}
\usepackage[colorlinks,
            linkcolor=red,
            anchorcolor=black,
            citecolor=hotpink]{hyperref}
\usepackage[accsupp]{axessibility}  % Improves PDF readability for those with disabilities.

\begin{document}
% \linenumbers
\pagestyle{headings}
\mainmatter
\def\ECCVSubNumber{3428}  % Insert your submission number here

\title{ScalableViT: Rethinking the Context-oriented Generalization of Vision Transformer} % Replace with your title

% \begin{comment}
% \titlerunning{ScalableViT} 
% \authorrunning{Rui Yang, Hailong Ma, Jie Wu, et al.} 
% \author{Rui Yang\inst{1}\thanks{Equal contribution.},\quad  Hailong Ma$^{2\, \star}$,\quad  Jie Wu$^{2\, \star \star}$,\quad Yansong Tang$^{1}$ ,\quad Xuefeng Xiao$^{2}$,\quad  Min Zheng$^{2}$,\quad  Xiu Li$^{1}$\thanks{Corresponding authors.}}
% % \institute{
% % $^1$Tsinghua Shenzhen International Graduate School, Tsinghua University\\
% % $^2$ByteDance Inc.\\}
% \institute{Tsinghua Shenzhen International Graduate School, Tsinghua University
% \email{\{r-yang20@mails,tang.yansong@sz,li.xiu@sz\}.tsinghua.edu.cn}\\
% \and
% ByteDance Inc.\\
% \email{\{mahailong.1206,wujie.10,xiaoxuefeng.ailab,zhengmin.666\}@bytedance.com}}

% \end{comment}
%******************
% \renewcommand{\thefootnote}{\fnsymbol{footnote}}
% \footnotetext[1]{These authors contributed equally to this work.}
% \footnotetext[2]{Corresponding authors.}

% \begin{comment}

% \end{comment}

%\begin{comment}
% \titlerunning{ECCV-22 submission ID \ECCVSubNumber} 
% \authorrunning{ECCV-22 submission ID \ECCVSubNumber} 
% \author{Anonymous ECCV submission}
% \institute{Paper ID \ECCVSubNumber}
%\end{comment}

% \begin{comment}
\titlerunning{ScalableViT} 
\authorrunning{Rui Yang, Hailong Ma, Jie Wu, et al.} 
\author{Rui Yang\inst{1}$^{\dagger \star}$,\quad  Hailong Ma\inst{2}$^\dagger$,\quad  Jie Wu\inst{2}$^\ddagger$,\quad Yansong Tang\inst{1},\quad Xuefeng Xiao\inst{2},\quad  Min Zheng\inst{2},\quad  Xiu Li\inst{1}$^\ddagger$}
\institute{Tsinghua Shenzhen International Graduate School, Tsinghua University
\email{\{r-yang20@mails,tang.yansong@sz,li.xiu@sz\}.tsinghua.edu.cn}\\
\and
ByteDance Inc.\\
\email{\{mahailong.1206,wujie.10,xiaoxuefeng.ailab,zhengmin.666\}@bytedance.com}}

\renewcommand{\thefootnote}{}
\footnotetext{$^\dagger$Equal contribution. $^\ddagger$Corresponding author. $^\star$This work was partly done while Rui Yang interned at ByteDance. Code: \url{https://github.com/Yangr116/ScalableViT}}

% \end{comment}
%******************
\maketitle

\begin{abstract}
The vanilla self-attention mechanism inherently relies on pre-defined and steadfast computational dimensions. Such inflexibility restricts it from possessing context-oriented generalization that can bring more contextual cues and global representations. To mitigate this issue, we propose a Scalable Self-Attention (SSA) mechanism that leverages two scaling factors to release dimensions of $query$, $key$, and $value$ matrices while unbinding them with the input. This scalability fetches context-oriented generalization and enhances object sensitivity, which pushes the whole network into a more effective trade-off state between accuracy and cost. Furthermore, we propose an Interactive Window-based Self-Attention (IWSA), which establishes interaction between non-overlapping regions by re-merging independent $value$ tokens and aggregating spatial information from adjacent windows. By stacking the SSA and IWSA alternately, the \textbf{Scalable} \textbf{Vi}sion \textbf{T}ransformer (ScalableViT) achieves state-of-the-art performance on general-purpose vision tasks. For example, ScalableViT-S outperforms Twins-SVT-S by \textbf{1.4\%} and Swin-T by \textbf{1.8\%} on ImageNet-1K classification.

\end{abstract}

\section{Introduction}

%-------------- figure visualize featmap
\begin{figure}
  \centering
  \includegraphics[width=\linewidth]{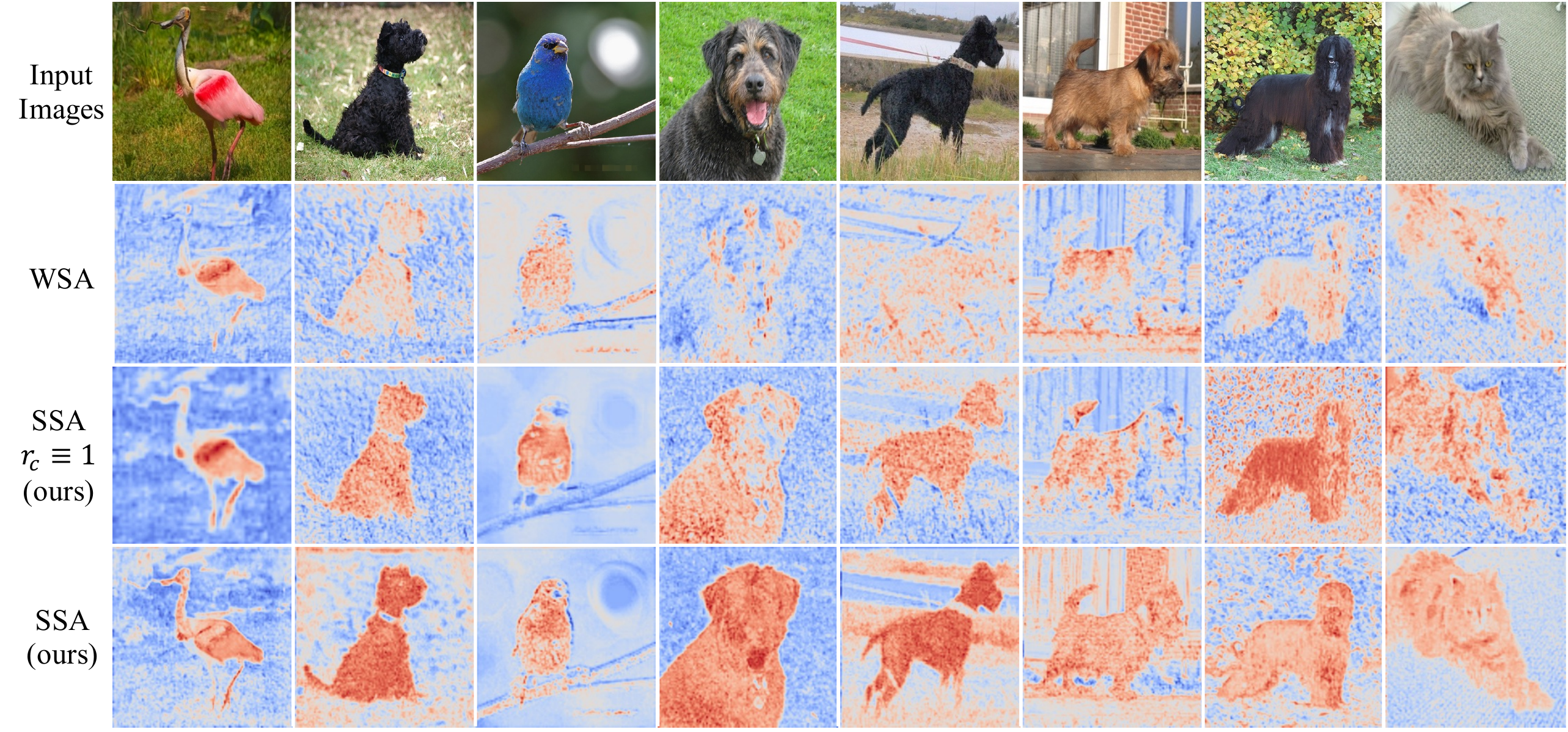}
   \caption{Visualization for feature maps in the Vision Transformer. We show the feature maps after the second Transformer blocks in Window-based Self-Attention (WSA)~\cite{Twins,swin} and Scalable Self-Attention (SSA). The activation from WSA is discontinuous because of a limited mapping dimension. SSA ($r_c\equiv1$) reduces computational overhead while retaining a global perception, ensuring its feature map is nearly continuous. $r_c\equiv1$ denotes no scaling factors in the channel dimension. SSA introduces scale factors to spatial and channel dimensions, modeling a holistic representation and a context-oriented generalization.}
   \label{figure:featmap}
\end{figure}
%-------------- figure visualize featmap

Convolutional Neural Networks (CNNs) dominated the computer vision field last few years, which attributes to their capacity in modeling realistic images from a local to global perception.
Although they have been widely applied in various vision tasks, there are still deficiencies in global visual perception. 
This global view is essential for downstream tasks, such as object detection and semantic segmentation. 
Recently, ViT~\cite{ViT} and its follow-ups~\cite{Deit,swin,Twins,crossformer} employed transformer encoders to address the image task and achieved comparable performance against their CNN counterparts because of the global receptive field.
However, the global perception of the Transformer entails an unaffordable computation since self-attention (the primary operation of the Transformer) is quadratically computed on the whole sequence.
% and its computation complexity is quadratic with the sequence length ($H\times W$ for the image).
To alleviate this overhead, typical Swin transformer~\cite{swin} employed Window-based Self-Attention (WSA), which partitioned a feature map into many non-overlapped sub-regions and enabled it to process large-scale images with linear complexity. They also proposed a novel Shifted Window-based Self-Attention (SWSA) to compensate for losses of potential long-range dependency. 
Twins~\cite{Twins} combined the WSA with Global Sub-sampled Attention (GSA) for better performance.

To gain an insight into the WSA~\cite{Twins,swin}, we visualize feature maps after the second block. As shown in Fig.~\ref{figure:featmap}, features captured by the WSA are dispersed, and their responses incline to partial rather than object-oriented.
% It may attribute to a fact that the dimension is fixed to $N$ invariably, which results in limited learning ability and the final performance being highly determined by the difficulty of input data.
It may attribute to an invariably fixed dimension that results in limited learning ability, thereby the final performance of the model being highly determined by the difficulty of input data.
To alleviate this problem, we develop a novel self-attention mechanism, termed Scalable Self-Attention (SSA), which simultaneously introduces two scaling factors ($r_n$ and $r_c$) to spatial and channel dimensions. Namely, SSA selectively applies these factors to $query$, $key$, and $value$ matrices ($Q$, $K$, and $V$), ensuring the dimension is more elastic and no longer deeply bound by the input. 
% On the one hand, SSA aggregates redundant tokens with similar semantic information to a more compact one via spatial scalability. In this way, unnecessary intermediate multiplication operations are eliminated and the computational complexity can be reduced significantly. 
On the one hand, SSA aggregates redundant tokens with similar semantic information to a more compact one via spatial scalability. Consequently, unnecessary intermediate multiplication operations are eliminated, and the computational complexity is reduced significantly.
% In the second row of Fig.~\ref{figure:featmap}, we can easily observe that the spatial scalability can realize nearly contiguous visual modeling for objects. However, it still losses some contextual cues.
In the third row of Fig.~\ref{figure:featmap}, we can easily observe that spatial scalability can bring nearly contiguous visual modeling for objects, but some contextual cues are still lost.
Hence, on the other hand, we expand the channel dimension to learn a more graphic representation.
% As depicted in the last row of Fig.~\ref{figure:featmap}, SSA successes to obtain complete object activation while maintaining context-oriented generalization via channel scalability. For instance, contextual cues of the cat in the last column are represented in detail.
As depicted in the last row of Fig.~\ref{figure:featmap}, SSA successfully obtains complete object activation while maintaining context-oriented generalization via channel scalability. For instance, the contextual cues of the cat in the last column are represented in detail.
% Furthermore, such scaling factors restore the output dimension craftily to $\mathbb{R}{^{N \times C}}$, which ensures the dimension aligns with the subsequent FFN layers and makes the residual connection feasible.
Such scaling factors also restore the output dimension to align with the input, which makes the residual connection feasible.

Moreover, we propose an Interactive Window-based Self-Attention (IWSA) that consists of a regular WSA and a local interactive module (LIM). 
The IWSA establishes information connections by re-merging independent $value$ tokens and aggregating spatial information from adjacent windows. Therefore, it no longer limits the self-attention to local windows, particularly non-overlapping windows. 
% Such characteristic enhances the desired global receptive field and take good advantage of the most significant superiority of the Transformer in a single layer.
Such characteristic enhances the desired global receptive field and takes good advantage of the most significant superiority of the Transformer in a single layer.
% The effectiveness of LIM for window-based self-attention is validated through an ablation study (see Tab.\ref{table:result_ablation_2}).
The effectiveness of LIM for WSA is validated in Tab.~\ref{table:result_ablation_2}.
To achieve a more efficient backbone for general vision tasks, we adopt a hierarchical design~\cite{ResNet,VGG} and propose a new Vision Transformer architecture, termed \textbf{ScalableViT}, which alternately arranges IWSA and SSA blocks in each stage. 
% Because of the context-oriented generalization and the global field achieved in the single block, it is more suitable for visual tasks.
% Because of the context-oriented generalization and the global field achieved in every block, it is more suitable for visual tasks.

Main contributions of our ScalableViT lie in two aspects:
\begin{itemize}

\item[$\bullet$] For the global self-attention, we propose SSA to supply context-oriented generalization in the vanilla self-attention block, which significantly reduces computational overhead without sacrificing contextual expressiveness.
% It significantly reduces the computational overhead while retaining the global perspective in self-attention. 
\item[$\bullet$] For the local self-attention, we design LIM to enhance the global perception ability of WSA.

\end{itemize}
% Both of them can model long-range dependency in a single layer instead of stacking more self-attention layer.
Both SSA and IWSA can model long-range dependency in a single layer instead of stacking more self-attention layers; hence, the ScalableViT is more suitable for visual tasks.
We employ ScalableViT on several vision tasks, including image-level classification on ImageNet~\cite{ImageNet}, pixel-level object detection and instance segmentation on COCO~\cite{COCO}, and semantic segmentation on ADE20K~\cite{ADE20K}. Extensive experiments demonstrate that the ScalableViT outperforms other state-of-the-art Vision Transformers with similar or less computational cost.
For example, ScalableViT-S achieves $\textbf{+1.4\%}$ gains against Twins-SVT-S and $\textbf{+1.8\%}$ gains against Swin-T on ImageNet-1K classification.
% For example, ScalableViT-S outperforms Twins-SVT-S by \textbf{1.4\%} and Swin-T by \textbf{1.8\%} on ImageNet-1K classification.

\section{Related Work}
The Transformer architecture~\cite{AttentionIsAllYouNeed} has become a common template for natural language processing (NLP) tasks due to its solid global modeling capabilities and convenient parallelization ability.
Inspired by this, many researchers tried to equip CNNs with the self-attention to modulate and augment outputs of convolutions~\cite{GCNet,DistentangledNonLocalNeuralNetworks,xia2022trt}. 
DETR~\cite{DETR} employed the self-attention mechanism to model relations between objects for end-to-end detection.
Others~\cite{CCNet,AACN,NonLocalNeuralNetworks} combined self-attention with convolutions for full-image contextual information.
% At the same time, the stand-alone self-attention network~\cite{SAN,StandAloneSelfAttention,Axial-DeepLa} proved that stacking attention layers alone can work for different vision tasks well.
Recently, the emergence of ViT~\cite{ViT}, DeiT~\cite{Deit}, and a series of follow-ups~\cite{swin,Cswin,T2T,PVT,CvT,Ceit,Twins} proved the bright prospect of the Vision Transformer.

%-------------------------------------------------------------------------
\subsection{Vision Transformer}
ViT~\cite{ViT} applied standard Transformer encoders to build a convolution-free image classifier by decomposing the image into a sequence of non-overlapping patches directly. Although it harvested promising results, a gap still existed between data-hungry Transformers and top-performing CNNs~\cite{EfficientNet} when only training on the midsize ImageNet-1K~\cite{ImageNet} from scratch. In order to bridge this gap, DeiT~\cite{Deit} proposed a token-based distillation procedure and a data-efficient training strategy to optimize the Transformer effectively. Later, the follow-ups improved different aspects of the ViT, making them more suitable for vision tasks. T2T-ViT~\cite{T2T} optimized the tokenization by concatenating the neighboring tokens into one token. DynamicViT~\cite{DynamicViT} pruned the tokens of less importance in a dynamic way for a better lightweight module. Cvt~\cite{CvT}, CeiT~\cite{Ceit} incorporated the convolution designs into the self-attention or the FFN to enhance the locality.
CPVT~\cite{CPVT} utilized the implicit position representation ability from convolutions (with zero padding) to encode the conditional position information for inputs with the arbitrary size.
Then, hierarchical pyramid structures~\cite{PVT,swin,Cswin,Twins} were performed by progressively shrinking the number of tokens and replacing the class token with the average pooling. Thus, the Transformer, supported by multi-level features~\cite{FPN}, can handle object detection and image segmentation tasks conveniently. In this paper, we develop a Vision Transformer, ScalableViT, which achieves a better accuracy and cost trade-off on visual tasks.

%-------------------------------------------------------------------------
\subsection{Local Self-Attention}
The computational complexity of the self-attention mechanism is a barrier that confines it in only downsampled feature maps or small images. Thus, several previous studies~\cite{StandAloneSelfAttention,LocalRelationNetworks,Axial-DeepLa,CCNet,InterlacedSparseSelfAttention} proposed decomposing the global self-attention into much paralleled local self-attention to handle expensive computation burdens. However, this local self-attention limits the receptive field that is critical to dense predict tasks. \cite{CCNet,Axial-DeepLa} proposed generating the sparse attention map on a criss-cross path to realize global interaction. \cite{InterlacedSparseSelfAttention} captured the information from all the other positions via interlacing elements between different local windows. HaloNet~\cite{Halonet} used the overlapped local windows to add the interactions between independent windows. After ViT~\cite{ViT} showed competitiveness, several follow-ups~\cite{swin,Twins,ShuffleTransformer,HRFormer} applied the self-attention within non-overlapped local windows for linear computational complexity. To compensate for lost information, Swin Transformer~\cite{swin} introduced a novel shifted window strategy, and Twins Transformer~\cite{Twins} baked sparse global attention~\cite{PVT} after WSA. We design the IWSA, which can aggregate information from a collection of discrete $value$ tokens and enable local self-attention to model long-range dependency in a single block.

\section{Method}
In this section, we elaborately introduce the architecture of ScalableViT and mainly focus on SSA and IWSA mechanisms. SSA simultaneously introduces different scale factors into spatial and channel dimensions to maintain context-oriented generalization while reducing computational overhead. IWSA enhances the receptive field of local self-attention by aggregating information from a set of discrete $value$ tokens. Both have linear computational complexity and can learn long-range dependency in a single layer. 

%------------------------------------------------------------------------
%-------------- figure architecture
\begin{figure}[t] 
  \centering
  \includegraphics[width=\linewidth]{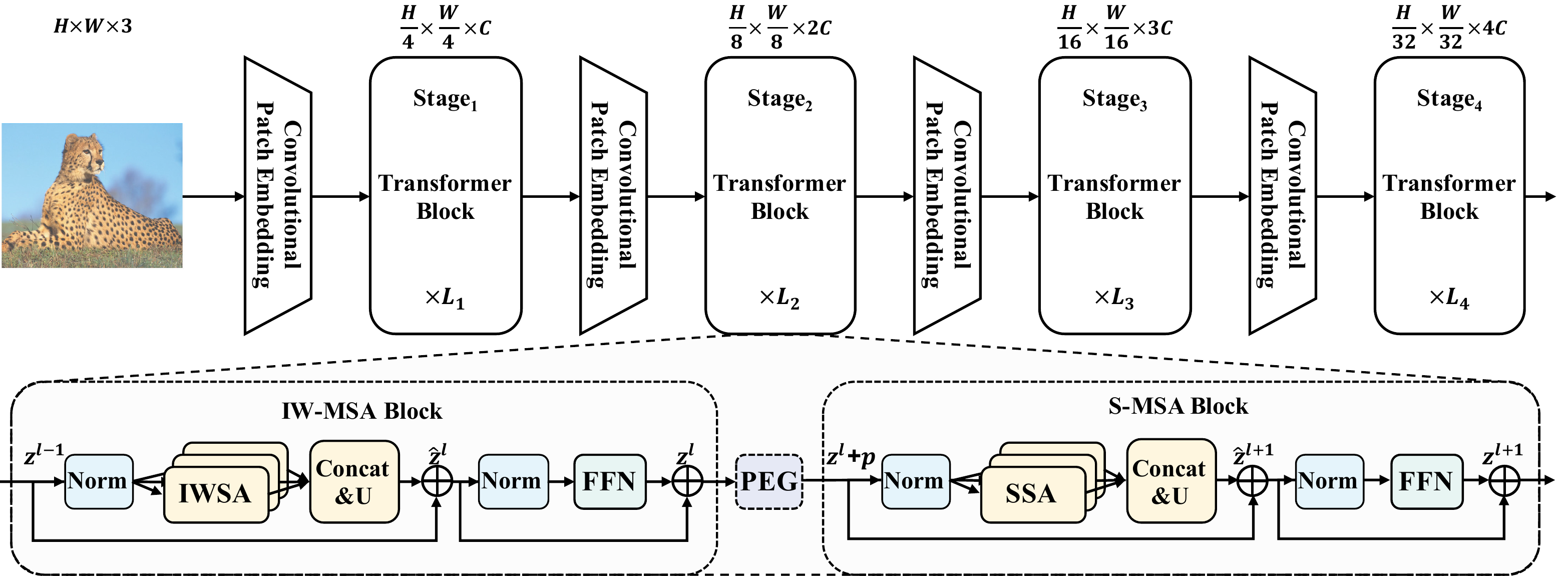}
   \caption{The architecture of the ScalableViT. IW-MSA and S-MSA, the multi-head format of IWSA and SSA, are organized alternately in each stage.  A PEG~\cite{CPVT} is placed between two blocks in the front of each stage to encode implicit position information dynamically.}
   \label{figure:Architecture}
\end{figure}
%-------------- figure architecture
\subsection{Overall Architecture}
The architecture of ScalableViT is illustrated in Fig.~\ref{figure:Architecture}. 
% For an input image with size $H \times W \times 3$,  a $7 \times 7$ convolution layer with stride 4  is regarded as patch embedding to obtain the initial patch tokens, and the channel dimension of each token is expanded to C. 
For an input image with size $H \times W \times 3$, a convolutional patch embedding layer ($7 \times 7$, stride $4$) is used to obtain a collection of tokens ($\frac{H}{4} \times \frac{W}{4}$) and project the channel dimension to $C$.
Then, these initial tokens will pass through four stages which contain a series of Transformer blocks. Between two adjacent stages, another convolutional patch embedding layer ($3 \times 3$, stride $2$) is utilized to merge tokens and double the channel dimension. For the $i^{th}$ stage, there are $\frac{H}{2^{i+1}}\times \frac{W}{2^{i+1}}$ input tokens with $2^{i-1}C$ channels and $L_i$ Transformer blocks. As a result, the quantity of tokens will eventually be reduced to $\frac{H}{32} \times \frac{W}{32}$. This architecture enables us to obtain a hierarchical representation similar to the typical backbones based on CNNs~\cite{ResNet,VGG}.
This merit allows ScalableViT to naturally migrate to various vision tasks, such as object detection and segmentation.
In each stage, we devise an alternate arrangement of IW-MSA and S-MSA blocks to organize the topological structure.
In the front of each stage, a position encoding generator (PEG)~\cite{CPVT} is inserted between two Transformer blocks to generate position embedding dynamically.

%-------------- figure SSA and LIM
\begin{figure}[t]
\begin{minipage}[t]{0.6\linewidth}
\centering
\includegraphics[width=1.0\linewidth]{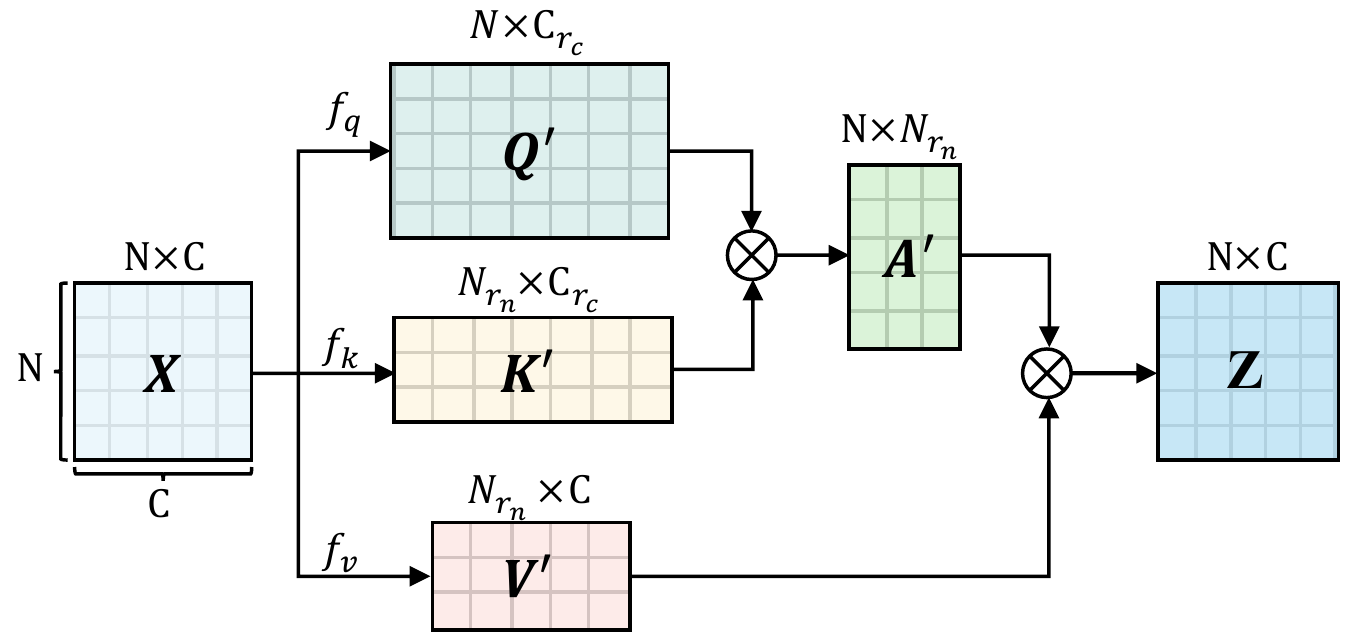}
\caption{The diagram of SSA. Two scale factors, $r_n$ and $r_c$, are introduced to the spatial and channel dimension for better computational efficiency and representational sufficiency.}
\label{figure:SSA}
\end{minipage}
\quad
\begin{minipage}[t]{0.35\linewidth}
\centering
\includegraphics[width=0.8\linewidth]{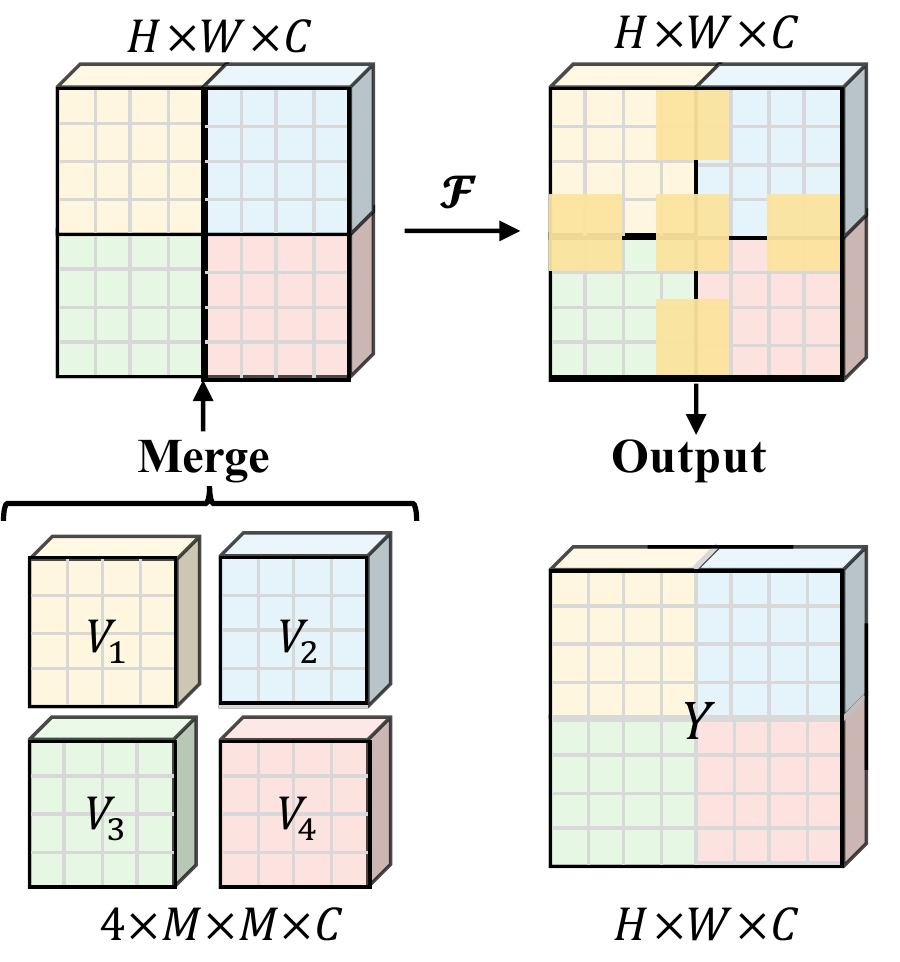}
\caption{The diagram of LIM. It merges a series of discrete $V_i$ by the function $\mathcal{F}$, and outputs an interacted $Y$.}
\label{figure:LIM}
\end{minipage}
\end{figure}
%-------------- figure SSA and LIM

%------------------------------------------------------------------------
\subsection{Scalable Self-Attention}
\label{subsec:ScalableAttention}

Self-attention is a critical mechanism in the Transformer, and the vanilla self-attention can be calculated as:
\begin{equation}
	Z = A(X) V(X) = Softmax(\frac{Q(X)K(X)^T}{\sqrt{d_k}}) V(X),
	\label{eq:attention}
\end{equation}
where $A(X)$ is the attention matrix of the input $X$; $Q(X), K(X), V(X) \in \mathbb{R}{^{N\times C}}$ are the $query$, $key$, and $value$ matrices; $d_k$ is the channel dimension of $query$ or $key$ matrix; $N$ is the number of tokens in each matrix, and $C$ is the channel dimension.
The original self-attention mechanism obtains a global receptive field by establishing associations between all input tokens, which is a vital advantage of the Transformer over CNNs. However, it has quadratic computational overhead with $N$, leading to inefficiency in the intermediate multiplication operations. 

Generally, there is much homologous information in natural images, but vanilla self-attention still calculates their similarity. Notably, not all information is necessary to calculate self-attention in the Vision Transformer. For example, similar background tokens should be aggregated as one representative token to attend to other foreground tokens.
Namely, the dimension of $Q(X)$, $K(X)$, and $V(X)$ should not be bounded with the input $X$. 
More importantly, the fixed dimension results in limited learning ability.
Thus, we develop the Scalable Self-Attention (SSA), where two scaling factors ($r_n$ and $r_c$) are introduced to spatial and channel dimensions, respectively, resulting a more efficient intermediate calculation than the vanilla one.
As illustrated in Fig.~\ref{figure:SSA}, the spatial dimension $N$ and channel dimension $C$ are selectively scaled to $N\times r_n$ and $C\times r_c$, respectively, by three transformation functions $f_q(\cdot)$, $f_k(\cdot)$, and $f_v(\cdot)$. 
These scaling factors can also restore the output dimension to align with the input, making the subsequent FFN layers and residual connections feasible.
% These scaling factors can restore the output dimension to align with input afresh so that the subsequent FFN layers and residual connections are feasible.
As a result, the intermediate dimension is more elastic and no longer deeply bound with the input $X$. The model can reap context-oriented generalization while dwindling computational overhead significantly.
SSA can be naturally written as:
\begin{gather}
	Z^{\prime} = A^{\prime}(X) V^{\prime}(X) = Softmax(\frac{Q^{\prime}(X){K^{\prime}(X)}^T}{\sqrt{d_k^{\prime}}}) V^{\prime}(X), \\
	Q^{\prime}(X) = f_q(X), \quad
	K^{\prime}(X) = f_k(X), \quad
	V^{\prime}(X) = f_v(V),
	\label{eq:Scalable_attention}
\end{gather}
where $Q^{\prime}(X) \in \mathbb{R}{^{N\times Cr_c}},  K^{\prime}(X) \in \mathbb{R}{^{Nr_n\times Cr_c}}, V^{\prime}(X) \in \mathbb{R}{^{Nr_n\times C}}$ are the scaled $query$, $key$ and $value$ matrices of the input $X\in{\mathbb{R}^{H\times W\times C}}$; $A^{\prime}(X) \in \mathbb{R}{^{N\times Nr_n}}$ is the scaled attention matrix; $Z^{\prime}$ is the weighted sum of $V^{\prime}(X)$.
The transformation $f_q(\cdot)$ scales the channel dimension of $query$ from $C$ to $C_{r_c}$.
$f_k(\cdot)$ is the scaling function for $key$, which scales the spatial dimension from $N$ to $N_{r_n}$ while scaling the channel dimension from $C$ to $C_{r_c}$.
$f_v(\cdot)$ is the scaling function for $value$, which scales the spatial dimension from $N$ to $N_{r_n}$.
% $f_q$ is the linear projection for $query$, which scales the channel dimension to $C_{r_c}$. 
% $f_k$ is the scaling function for $key$. It first scales the spatial dimension to $N_{r_n}$ through a convolutional layer and then linearly projects the channel dimension to $C_{r_c}$.
% $f_v$ is the scaling function for $value$, which only scales the number of tokens to $N_{r_n}$ through a convolutional layer in the spatial dimension.
Hence, some unnecessary intermediate multiplication is decreased significantly. The computation complexity of the proposed SSA is equal to $\mathcal{O}(N N_{r_n} C + N N_{r_n} C_{r_c})$ that is linear with the input size ($N = H \times W$).
% In practice, three transformations are operated by convolution and fused with the linear projection getting $query$, $key$, and $value$ for utility and briefness.
For utility and briefness, three transformations are operated by convolutions and cooperates with linear projections to get the scaled $query$, $key$, and $value$ matrices.
The efficient SSA does not change the size of $Z$ and can be expanded to the Scalable Multi-Head Self-Attention (S-MSA) easily. 

More importantly, the introduced spatial and channel scalability can bring context-oriented generalization (see Fig.~\ref{figure:featmap}). If only spatial scalability is introduced ($r_c \equiv 1$), there would realize nearly contiguous visual modeling for objects but a lack of critical graphic representation. When further introducing channel scalability, SSA can successfully maintain contextual cues and obtain complete object activation, which is essential in visual tasks. The values of these two scaling factors vary with model configurations and different network stages. As the network gradually deepens, the quantity of tokens shrinks, and the degree of redundancy is also dropped. Thus, $r_n$ is largen with the stage depth. 
% Similarly, the value of $r_c$ is different for networks magnitudes. The reason is that channel dimension mismatch does not always emerge in the self-attention calculation. 
% We found that the channel dimension mismatch between attention and MLP-layer is deficient rather than redundant for small configurations. 
Similarly, the channel dimension does not always mismatch with spatial dimension in the self-attention operation. 
Thus, we set $r_c \ge 1$ in ScalableViT-S and ScalableViT-B. Because of a too-large channel dimension, we set $r_c \le 1$ in ScalableViT-L.
% In ScalableViT-L, this mismatch is reflected in which the relatively large channel dimension required by MLP is redundant for attention calculation. Therefore, in larger models, we set $r_c < 1$ in some stages. 
Details about two scale factors are displayed in Table~\ref{table:configurations}.

%------------------------------------------------------------------------

\subsection{Interactive Window-based Self-Attention}
% Besides efficient self-attention, several earlier research has restricted self-attention to local windows, especially some non-overlapping windows, to avoid quadratic complexity concerning the number of tokens. 
Besides the efficient self-attention~\cite{PVT}, earlier researches have developed the local self-attention~\cite{swin,Twins} to avoid the quadratic computational complexity with the number of tokens.
% For example, assume that an image with size $H \times W$ and channel dimension $C$ could be divided into multiple partial windows containing $M\times M$ tokens. Moreover, the discrete $Q_i$, $K_i$, $V_i$ would be mapped out from the corresponding windows. 
For example, WSA divides an image ($H \times W\times C$) into multiple partial windows which contains $M\times M$ tokens.
Then, the self-attention would be calculated in every isolated window and produce a set of discrete outputs $\{Z_n\}_{n=1}^{\frac{H}{M} \times \frac{W}{M}}$, where $Z_n$ can be calculated as:
\begin{equation}
  Z_n = A_n(X_n) V_n(X_n)= Softmax(\frac{Q_n(X_n) K_n(X_n)^T}{\sqrt{d_k}})V_n(X_n),
  \label{eq:Local_attention}
\end{equation}
in which $X_n\in{\Omega_n^{M\times M\times C}}$ is the partial window field; $Q_n(X_n), K_n(X_n), V_n (X_n)\in \mathbb{R}{^{M^2 \times C}}$ are the $query$, $key$, and $value$ matrices of the discrete window $X_n$, respectively.
% $K_i \in \mathbb{R}{^{M^2 \times C}}$ is key of window $\mathcal{G}$, and $V_i \in \mathbb{R}{^{M^2 \times C}}$ is the value of window $\mathcal{G}$. 
$d_k$ is equal to the channel dimension of discrete $query/key$ tokens.
Finally, a collection of discrete $\{Z_n\}_{n=1}^{\frac{H}{M} \times \frac{W}{M}}$ is merged back to $Z  \in \mathbb{R}{^{N \times C}}$. 
Thus, for an image, the computational complexity of attention would be reduced from $\mathcal{O}(2 H^2 W^2 C)$ to $\mathcal{O}(2 M^2 H W C)$. The WSA can be suitable for various vision tasks that require high-resolution input due to its linear complexity.

% However, such computationally efficient WSA only yields an activation map with integrated shapes but isolated features (see Fig.~\ref{figure:featmap}), which ascribed to the missed global receptive field in a single layer. It is far from the initial aims of self-attention.
However, such computationally efficient WSA yields a feature map with an integrated shape but isolated activation (see Fig.~\ref{figure:featmap}), which ascribed to the missed global receptive field in a single layer. This is far from the initial aims of self-attention.
To alleviate above problem, we propose Interactive Window-based Self-Attention (IWSA) that incorporates a local interactive module (LIM) into WSA, as illustrated in Fig.~\ref{figure:LIM}.
After getting a collection of discrete $values$ $\{V_n(X_n)\}_{n=1}^{\frac{H}{M} \times \frac{W}{M}}$, the LIM reshapes them into $M \times M \times C$ and merges them into a shape-integrated $value$ map $V \in \mathbb{R}{^{H \times W \times C}}$.
% Then, a function $F$ is used to establish marriages and connections between adjacent $V_i$. So, the output is an integrated feature map with the global receptive field, which is reshaped and added with the discrete $Z_i$ to form the final output $Z_i^{\prime}$. Finally, a collection of $Z_i^{\prime}$ with interactive information is merged to $Z$ that has global information. In order to implement friendly, a depth-wise convolution with zero padding is used to take the place of function $F$. 
Subsequently, a function $\mathcal{F}(x)$ is employed to establish marriages and connections between adjacent $V_n(X_n)$s. As a result, the output $Y=\mathcal{F}(V)$ is an integrated feature map with global information. Finally, this feature map is added on $Z$ as the final output $Z^{\prime}$. 
% In order to implement friendly, a depth-wise convolution with zero padding is used to take the place of function $F$.
Without loss of generality, the IWSA is calculated as:
\begin{equation}
  Z^{\prime} = Z + \mathcal{F}(V),
  \label{eq:Local_interactive_attention}
\end{equation}
% \begin{align}
%   Z_i^{\prime} = Z_i + [Deepwise(V)]_i, \quad i\in{\mathcal{G}}
%   \label{eq:Local_interactive_attention}
% \end{align}
% where $[Deepwise(V)]_i \in \mathbb{R}{^{M^2 \times C}}$ is a locally reshaped result of $Deepwise(V) \in \mathbb{R}{^{H \times W \times C}}$ in $\mathcal{G}$ and $V$ is a reshaped $value$ map with integrated shape.
where $Z  \in \mathbb{R}{^{N \times C}}$ is merged by a set of $\{Z_n\}_{n=1}^{\frac{H}{M} \times \frac{W}{M}}$.
% and $F(V)$ is the output of the LIM, and $F(x)$ is a function for information interaction. 
In order to be implemented friendly, a depth-wise convolution with zero padding is employed to take the place of function $\mathcal{F}(x)$.
% If the depth-wise convolution is $k \times k$ in kernel size(set to 3 by default), the increase in the computational cost of the entire module is $O(k^2HWC)$, which is negligible in practice.
If the kernel size of this depth-wise convolution is $k \times k$ (set to 3 by default), the computational cost from the LIM is negligible in practice.
Additionally, \cite{IslamJB20} demonstrated that the convolution with zero padding could implicitly encode position information through experiments. Thus, IWSA allows self-attention to benefit from the translation invariance. 
Furthermore, IWSA can be easily expanded to Interactive Window-based Multi-head Self-Attention (IW-MSA) format easily if calculated in different heads. 
% Our IWSA can expand to local interactive multi-head self-attention(IW-MSA) easily if calculated it in different heads. 

% CoaT~\cite{CoaT} also introduced a depth-wise convolution into the self-attention. However, they only considered the convolution as a positional encoding method and bound. limited it in the discrete $V_n(X_n)$. 
% Differently, we regard our LIM as a matchmaker and apply it on the spliced $value$ map $V$. 
CoaT~\cite{CoaT} also introduced a depth-wise convolution into self-attention. However, they only considered the convolution as a positional encoding method and inserted it deeply into the calculation. If this convolution is expanded into the WSA, it would be limited in the discrete $V_n(X_n)$, which is denoted as local enhanced module (LEM). Differently, we regard our LIM as a matchmaker, which is applied on the spliced $value$ map $V$ and parallels with self-attention.
By making the sufficient ablation study in Section~\ref{sec:ablation_study}, we demonstrate that LIM is capable of delivering stable improvements, especially for downstream tasks.

%------------------------------------------------------------------------
\subsection{Position Encoding}
Besides the position information introduced by LIM, we utilize the positional encoding generator (PEG)~\cite{CPVT}, composed of a convolution layer with fixed weights, to acquire implicit positional information. As illustrated in Fig.~\ref{figure:Architecture}, it is plugged between two consecutive Transformer blocks, with only one in the front of each stage. 
After the PEG, input tokens are sent to subsequent blocks where position bias could enable the Transformer to realize the input permutation.

%------------------------------------------------------------------------
\subsection{Architecture Variants}
In order to fairly compare with other models under similar computation complexity, we set three models: ScalableViT-S, ScalableViT-B, and ScalableViT-L. The detailed configurations are provided in Table~\ref{table:configurations}, where $r_c$ and $r_n$ denote expansion or reduction factors for channel and spatial dimensions, respectively, as described in Section~\ref{subsec:ScalableAttention}. Due to the varying representational capability, we set different $r_c$ for three models. 
Additionally, the number of blocks, channels, and heads varies with the computational cost.

% --------------------------------- table configurations
\begin{table}[t]
\centering
\caption{Detailed configurations of ScalableViT series. $r_c$ and $r_n$ are scale factors for the channel and the spatial dimensions, respectively. '\#Blocks' and '\#Heads' refer to the number of blocks ($[L_1, L_2, L_3, L_4]$) and heads in four stages, respectively.  '\#Channels' refers to the channel dimension of the first stage.}
\label{table:configurations}
\resizebox{0.9\textwidth}{!}{
\begin{tabular}{@{}c|c|c|c|c|c@{}}
\toprule
Models        & \#Channels & \#Blocks       & \#Heads        & $r_c$                  & $r_n $                                                 \\ \midrule
ScalableViT-S & 64         & {[}2,2,20,2{]} & {[}2,4,8,16{]}  & {[}1.25,1.25,1.25,1.0{]} & {[}$\frac{1}{64}$, $\frac{1}{16}$, $\frac{1}{4}$, $1${]} \\
ScalableViT-B & 96         & {[}2,2,14,6{]} & {[}3,6,12,24{]} & {[}2.0,1.25,1.25,1.0{]}  & {[}$\frac{1}{64}$, $\frac{1}{16}$, $\frac{1}{4}$, $1${]} \\
ScalableViT-L & 128        & {[}2,6,12,4{]} & {[}4,8,16,32{]} & {[}0.25,0.5,1.0,1.0{]}   & {[}$\frac{1}{64}$, $\frac{1}{16}$, $\frac{1}{4}$, $1${]} \\ \bottomrule
\end{tabular}
}
% \end{center}
\end{table}
% --------------------------------- table configurations

\section{Experiments}
\label{sec:experiments}
In the following, we compare the proposed model with other state-of-the-art works on ImageNet-1K~\cite{ImageNet}, COCO~\cite{COCO}, and ADE20K~\cite{ADE20K}. Then, we conduct ablation studies on the upgraded parts to verify their effectiveness.

%-------------------------------------------------------------------------
\subsection{Image Classification on ImageNet-1K}
\label{sec:classification}
\noindent \textbf{Settings.}
Image classification experiments are conducted on the ImageNet-1K~\cite{ImageNet} dataset. All settings mainly follow DeiT~\cite{Deit}. 
During training, we apply data augmentation and regularization strategies in \cite{Deit}.
% , including random cropping, random horizontal flipping~\cite{RandomHorizontalFilpping}, mixup~\cite{Mixup}, CutMix~\cite{CutMix}, random erasing~\cite{RandomEarsing}, label-smoothing~\cite{LabelSmoothing}, and stochastic depth~\cite{StochasticDepth}. 
We employ the AdamW optimizer~\cite{AdamW} to train models for 300 epochs from scratch. The learning rate is set to 0.001 initially and varies with the cosine scheduler. The global batchsize is set to 1024 on 8 V100 GPUs. 
During testing on the validation set, the shorter side of an input image is first resized to 256, and a center crop of 224 × 224 is used to evaluate the classification accuracy.
% More details are supplied in the Appendix~\ref{sec:settings}. 

% \vspace*{1\baselineskip}
\noindent \textbf{Result.}
Classification results on ImageNet-1K are reported in Table~\ref{table:result_ImageNet}, where all models are divided into small (around 4G), base (around 9G), and large (around 15G) levels according to computation complexity (FLOPs). ScalableViT-S with a two-layer head outperforms comparable models ($1.4\%$ better than Twins-SVT-S, and $1.8\%$ better than Swin-T). Moreover, it can even approach or exceed other base models. For the base level, ScalableViT-B surpasses Twins-SVT-B by $0.9\%$ and SWin-S by $1.1\%$ with similar FLOPs. ScalableViT-L also achieves a prominent accuracy-cost trade-off. Additionally, our ScalableViT outperforms the EfficientNet by $0.2\%$, $0.5\%$, and $0.4\%$ under three magnitude receptively. 
% We discover that our ScalableViT is the only Transformer-based architecture that achieves better performance than EfficientNet at the same cost, which illustrates that it may be a viable alternative to the CNN-based model for classification.

% --------------------------------- table imagenet
\begin{table}[t]
\centering
\caption{Comparison with different state-of-the-art backbones on ImageNet-1K classification. Except for EfficientNet, other models are trained and evaluated on $224\times224$ input size. Top-1 refers to top-1 accuracy ($\%$).}
\label{table:result_ImageNet}
\begin{subtable}{0.47\textwidth}
\resizebox{\textwidth}{!}{
\begin{tabular}{@{}l|cc|c@{}}
        \toprule
        Method                               & \#Param. & FLOPs                     & Top-1         \\ \midrule
\multicolumn{4}{c}{ConvNet}                                                                                 \\ \midrule
\multicolumn{1}{l|}{RegNetY-4G~\cite{RegNet}}                   & 21M        & \multicolumn{1}{c|}{4.0G}  & 80.0          \\
\multicolumn{1}{l|}{RegNetY-8G~\cite{RegNet}}                   & 39M        & \multicolumn{1}{c|}{8.0G}  & 81.7          \\
\multicolumn{1}{l|}{RegNetY-16G~\cite{RegNet}}                  & 84M        & \multicolumn{1}{c|}{16.0G} & 82.9          \\
\multicolumn{1}{l|}{EfficientNet-B4~\cite{EfficientNet}}              & 19M        & \multicolumn{1}{c|}{4.2G}  & 82.9          \\
\multicolumn{1}{l|}{EfficientNet-B5~\cite{EfficientNet}}              & 30M        & \multicolumn{1}{c|}{9.9G}  & 83.6          \\
\multicolumn{1}{l|}{EfficientNet-B6~\cite{EfficientNet}}              & 43M        & \multicolumn{1}{c|}{19.0G} & 84.0          \\\midrule
\multicolumn{4}{c}{Transformer}                                                                             \\ \midrule
\multicolumn{1}{l|}{DeiT-Small/16~\cite{Deit}}   & 22M      & \multicolumn{1}{c|}{4.6G} & 79.9          \\
\multicolumn{1}{l|}{T2T-ViT-14~\cite{T2T}}      & 22M      & \multicolumn{1}{c|}{5.2G} & 81.5          \\
\multicolumn{1}{l|}{TNT-S~\cite{TNT}}           & 24M      & \multicolumn{1}{c|}{5.2G} & 81.3          \\
\multicolumn{1}{l|}{CoaT-Lite(S)~\cite{CoaT}} & 20M      & \multicolumn{1}{c|}{4.0G} & 81.9          \\
\multicolumn{1}{l|}{PVT-Small~\cite{PVT}}       & 25M      & \multicolumn{1}{c|}{3.8G} & 79.8          \\
\multicolumn{1}{l|}{Swin-T~\cite{swin}}          & 29M      & \multicolumn{1}{c|}{4.5G} & 81.3          \\
\multicolumn{1}{l|}{CvT-13~\cite{CvT}}          & 20M      & \multicolumn{1}{c|}{4.5G} & 81.6          \\
\multicolumn{1}{l|}{Twins-SVT-S~\cite{Twins}}     & 24M      & \multicolumn{1}{c|}{2.9G} & 81.7          \\
\multicolumn{1}{l|}{CrossFormer-S~\cite{crossformer}}   & 31M      & \multicolumn{1}{c|}{4.9G} & 82.5          \\
 \multicolumn{1}{l|}{\textbf{ScalableViT-S(ours)}} & 32M      & \multicolumn{1}{c|}{4.2G}                      & \textbf{83.1} \\ \bottomrule
\end{tabular}
}
\end{subtable}
% base
\begin{subtable}{0.48\textwidth}
\resizebox{\textwidth}{!}{
\begin{tabular}{@{}l|cc|c@{}}
\toprule
Method                 & \#Param.             & FLOPs                 & Top-1                \\ \midrule
\multicolumn{4}{c}{Transformer}                                                                             \\ \midrule
T2T-ViT-19~\cite{T2T}             & 39M                  & 8.9G                  & 81.9                 \\
CoaT(S)~\cite{CoaT}       & 22M                  & 12.6G                  & 82.1                 \\
CoaT-Lite(M)~\cite{CoaT}       & 45M                  & 9.8G                  & 83.6                 \\
PVT-Medium~\cite{PVT}             & 44M                  & 6.7G                  & 81.2                 \\
Swin-S~\cite{swin}                 & 50M                  & 8.7G                  & 83.0                 \\
CvT-21~\cite{CvT}                 & 32M                  & 7.1G                  & 82.5                 \\
Twins-SVT-B~\cite{Twins}            & 56M                  & 8.6G                  & 83.2                 \\
CrossFormer-B~\cite{crossformer}          & 52M                  & 9.2G                  & 83.4                 \\
\textbf{ScalableViT-B(ours)} & 81M                  & 8.6G                  & \textbf{84.1}        \\
                       \midrule
Deit-Base/16~\cite{Deit}                 & 86M      & 17.6G & 81.8          \\
T2T-ViT-24~\cite{T2T}                   & 64M      & 14.1G & 82.3          \\
TNT-B~\cite{TNT}                        & 66M      & 14.1G & 82.8          \\
PVT-Large~\cite{PVT}                    & 61M      & 9.8G  & 81.7          \\
Swin-B~\cite{swin}                       & 88M      & 15.4G & 83.3          \\
Twins-SVT-L~\cite{Twins}                  & 99M      & 15.1G & 83.7          \\
CrossFormer-L~\cite{crossformer}                & 92M      & 16.1G & 84.0          \\
\textbf{ScalableViT-L(ours)} & 104M     & 14.7G & \textbf{84.4} \\ \bottomrule
\end{tabular}
}
\end{subtable}
\end{table}
% --------------------------------- table imagenet

%-------------------------------------------------------------------------
\subsection{Object Detection on COCO}
\label{subsec: detection results}
% --------------------------------- table RetianaNet
\begin{table}[t]
\caption{Results on COCO object detection using the RetinaNet~\cite{RetinaNet} and Mask R-CNN~\cite{MaskRCNN} framework. 1$\times$ refers to 12 epochs, and 3$\times$ refers to 36 epochs. MS means multi-scale training. AP$^b$ and AP$^m$ denotes box mAP and mask mAP, respectively. FLOPs are measured at resolution $800 \times 1280$.}
\label{table:result_COCO}
\begin{subtable}{\textwidth}
\resizebox{\textwidth}{!}{
\begin{tabular}{@{}l|cc|cccccc|cccccc@{}}
\toprule
\multirow{2}{*}{Backbone}    & \#Param.      & FLOPs      & \multicolumn{6}{c|}{RetinaNet 1$\times$}      & \multicolumn{6}{c}{RetinaNet 3$\times$ + MS}  \\ 
\cmidrule(l){2-15}  & (M)        & (G)        & AP$^b$            & AP$_{50}^b$           & AP$_{75}^b$           & AP$_S^b$           & AP$_M^b$            & \multicolumn{1}{c|}{AP$_L^b$}            & AP$^b$            & AP$_{50}^b$            & AP$_{75}^b$            & AP$_S^b$            & AP$_M^b$           & AP$_L^b$            \\ \midrule
ResNet50~\cite{ResNet}                     & 38       & 239        & 36.3          & 55.3          & 38.6          & 19.3          & 40.0          & \multicolumn{1}{c|}{48.8}          & 39.0          & 58.4          & 41.8          & 22.4          & 42.8          & 51.6          \\
PVT-Small~\cite{PVT}                    & 34       & 226        & 40.4          & 61.3          & 43.0          & 25.0          & 42.9          & \multicolumn{1}{c|}{55.7}          & 42.2          & 62.7          & 45.0          & 26.2          & 45.2          & 57.2          \\
Swin-T~\cite{swin}                       & 39       & 245        & 41.5          & 62.1          & 44.2          & 25.1          & 44.9          & \multicolumn{1}{c|}{55.5}          & 43.9          & 64.8          & 47.1          & 28.4          & 47.2          & 57.8          \\
Twins-SVT-S~\cite{Twins}                  & 34       & 210        & 43.0          & 64.2          & 46.3          & 28.0          & 46.4          & \multicolumn{1}{c|}{57.5}          & 45.6          & 67.1          & 48.6          & 29.8          & 49.3          & 60.0          \\
CrossFormer-S~\cite{crossformer}                  & 41       & 272        & 44.4          & 65.8          & 47.4          & 28.2          & 48.4          & \multicolumn{1}{c|}{59.4}          & ---          & ---          & ---          & ---          & ---          & ---          \\
\textbf{ScalableViT-S(ours)} & 36 & 238 & \textbf{45.2} & \textbf{66.5} & \textbf{48.4} & \textbf{29.2} & \textbf{49.1} & \multicolumn{1}{c|}{\textbf{60.3}} & \textbf{47.8} & \textbf{69.2} & \textbf{51.2} & \textbf{31.4} & \textbf{51.5} & \textbf{63.4} \\ \midrule
ResNet101~\cite{ResNet}                    & 58       & 315        & 38.5          & 57.8          & 41.2          & 21.4          & 42.6          & \multicolumn{1}{c|}{51.1}          & 40.9          & 60.1          & 44.0          & 23.7          & 45.0          & 53.8          \\
PVT-Medium~\cite{PVT}                   & 54       & 283        & 41.9          & 63.1          & 44.3          & 25.0          & 44.9          & \multicolumn{1}{c|}{57.6}          & 43.2          & 63.8          & 46.1          & 27.3          & 46.3          & 58.9          \\
Swin-S~\cite{swin}            & 60       & 335        & 44.5          & 65.7          & 47.5          & 27.4          & 48.0          & \multicolumn{1}{c|}{59.9}          & 46.3          & 67.4          & 49.8          & 31.1          & 50.3          & 60.9          \\
Twins-SVT-B~\cite{Twins}  & 67       & 326        & 45.3          & 66.7          & 48.1          & 28.5          & 48.9          & \multicolumn{1}{c|}{60.6}          & 46.9          & 68.0          & 50.2          & 31.7          & 50.3          & 61.8          \\
CrossFormer-B~\cite{crossformer}  & 62       & 389        & \textbf{46.2}          & \textbf{67.8}          & \textbf{49.5}          & \textbf{30.1}          & \textbf{49.9}          & \multicolumn{1}{c|}{\textbf{61.8}}          & ---          & ---          & ---          & ---          & ---          & ---          \\
\textbf{ScalableViT-B(ours)} & 85 & 330 & 45.8 & 67.3 & 49.2 & 29.9 & 49.5 & \multicolumn{1}{c|}{61.0} & \textbf{48.0} & \textbf{69.3} & \textbf{51.4} & \textbf{32.8} & \textbf{51.6} & \textbf{62.4} \\ \bottomrule
\end{tabular}}
\end{subtable}

\begin{subtable}{\textwidth}
\resizebox{\textwidth}{!}{
\begin{tabular}{@{}l|cc|cccccc|cccccc@{}}
\toprule
\multirow{2}{*}{Backbone}    & \#Param. & FLOPs & \multicolumn{6}{c|}{Mask R-CNN 1$\times$}  & \multicolumn{6}{c}{Mask R-CNN 3$\times$ + MS} \\ \cmidrule(l){2-15} 
                             & (M)   & (G)   & AP$^b$   & AP$_{50}^b$   & \multicolumn{1}{c|}{AP$_{75}^b$ }   & AP$^m$    & AP$_{50}^m$    & AP$_{75}^m$    & AP$^b$   & AP$_{50}^b$   & \multicolumn{1}{c|}{AP$_{75}^b$}   & AP$^m$   & AP$_{50}^m$   & AP$_{75}^m$           \\ \midrule
ResNet50~\cite{ResNet}                     & 44  & 260   & 38.0          & 58.6          & \multicolumn{1}{c|}{41.4}          & 34.4          & 55.1          & 36.7          & 41.0          & 61.7          & \multicolumn{1}{c|}{44.9}          & 37.1          & 58.4          & 40.1          \\
PVT-Small~\cite{PVT}                    & 44  & 245   & 40.4          & 62.9          & \multicolumn{1}{c|}{43.8}          & 37.8          & 60.1          & 40.3          & 43.0          & 65.3          & \multicolumn{1}{c|}{46.9}          & 39.9          & 62.5          & 42.8          \\
Swin-T~\cite{swin}                       & 48  & 264   & 42.2          & 64.4          & \multicolumn{1}{c|}{46.2}          & 39.1          & 64.6          & 42.0          & 46.0          & 68.2          & \multicolumn{1}{c|}{50.2}          & 41.6          & 65.1          & 44.8          \\
Twins-SVT-S~\cite{Twins}                  & 44  & 228   & 43.4          & 66.0          & \multicolumn{1}{c|}{47.3}          & 40.3          & 63.2          & 43.4          & 46.8          & 69.2          & \multicolumn{1}{c|}{51.2}          & 42.6          & 66.3          & 45.8          \\
CoaT-Lite(S)~\cite{CoaT}                  & 40  & ---   & 45.2          & ---          & \multicolumn{1}{c|}{---}          & 40.7          & ---          & ---          & 45.7          & ---          & \multicolumn{1}{c|}{---}          & 41.1          & ---          & ---          \\
CrossFormer-S~\cite{crossformer}                      & 50  & 301   & 45.4 & \textbf{68.0} & \multicolumn{1}{c|}{49.7} & 41.4 & \textbf{64.8} & 44.6 & --- & --- & \multicolumn{1}{c|}{---} & --- & --- & ---          \\
\textbf{ScalableViT-S(ours)} & 46     & 256     & \textbf{45.8}          & 67.6          & \multicolumn{1}{c|}{\textbf{50.0}}          & \textbf{41.7}          & 64.7          & \textbf{44.8}          & \textbf{48.7}          & \textbf{70.1}          & \multicolumn{1}{c|}{\textbf{53.6}}          & \textbf{43.6} & \textbf{67.2}          & \textbf{47.2} \\ \midrule
ResNet101~\cite{ResNet}                    & 63  & 336   & 40.4          & 61.1          & \multicolumn{1}{c|}{44.2}          & 36.4          & 57.7          & 38.8          & 42.8          & 63.2          & \multicolumn{1}{c|}{47.1}          & 38.5          & 60.1          & 41.3          \\
PVT-Medium~\cite{PVT}                   & 64  & 302   & 42.0          & 64.4          & \multicolumn{1}{c|}{45.6}          & 39.0          & 61.6          & 42.1          & 44.2          & 66.0          & \multicolumn{1}{c|}{48.2}          & 40.5          & 63.1          & 43.5          \\
Swin-S~\cite{swin}                       & 69  & 354   & 44.8          & 66.6          & \multicolumn{1}{c|}{48.9}          & 40.9          & 63.4          & 44.2          & 48.5          & 70.2          & \multicolumn{1}{c|}{53.5}          & 43.3          & 67.3          & 46.6          \\
Twins-SVT-B~\cite{Twins}                  & 76  & 340   & 45.2          & 67.6          & \multicolumn{1}{c|}{49.3}          & 41.5          & 64.5          & 44.8          & 48.0          & 69.5          & \multicolumn{1}{c|}{52.7}          & 43.0          & 66.8          & 46.6          \\
CoaT(S)~\cite{CoaT}                  & 42  & ---   & 46.5          & ---          & \multicolumn{1}{c|}{---}          & 41.8          & ---          & ---          & \textbf{49.0}          & ---          & \multicolumn{1}{c|}{---}          & 43.7          & ---          & ---          \\
CrossFormer-B~\cite{crossformer}                      & 72  & 408   & \textbf{47.2} & \textbf{69.9} & \multicolumn{1}{c|}{\textbf{51.8}} & \textbf{42.7} & \textbf{66.6} & \textbf{46.2} & --- & --- & \multicolumn{1}{c|}{---} & --- & --- & --- \\
\textbf{ScalableViT-B(ours)} & 95     & 349     & 46.8          & 68.7          & \multicolumn{1}{c|}{51.5}          & 42.5          & 65.8          & 45.9          & \textbf{49.0}          & \textbf{70.3}          & \multicolumn{1}{c|}{\textbf{53.6}}          & \textbf{43.8}          & \textbf{67.4}          & \textbf{47.5}          \\ \bottomrule
\end{tabular}}
\end{subtable}

\end{table}

\noindent \textbf{Settings.}
Object detection experiments are conducted on COCO 2017~\cite{COCO} dataset. We verify the model effectiveness on RetinaNet~\cite{RetinaNet} and Mask R-CNN~\cite{MaskRCNN} detection frameworks using the MMDetection~\cite{mmdetection}. Before training, we initialize the backbone with the weight pre-trained on ImageNet-1K, FPN with Xavier~\cite{Xavier} scheme, and other new layers with Normal scheme ($std=0.01$). All models utilize the same settings as \cite{Twins}: AdamW~\cite{AdamW} optimizer, $1\times$ (12 epochs), and $3\times$ (36 epochs) schedules with a global batchsize of 16 on 8 GPUs. 
For the $1 \times$ schedule, the short side of images is resized to 800 pixels, and the long side is never more than 1333 pixels. The learning rate is declined at the 8th and 11th epoch with a decay rate of $0.1$.
For the $3\times$ schedule, we adopt the multi-scale training, which randomly resizes the short side of images within the range of [480, 800] while keeping the longer side at most 1333. The learning rate is declined at the 27th and 33rd with a decay rate of $0.1$. 
% More details are provided in the Appendix~\ref{sec:settings}.  
% : AdamW~\cite{AdamW} optimizer, $1\times$(12epoch), and $3\times$(36epoch) schedule with a global batch size of 16 on 8 GPUs. For the $1 \times$ schedule, the short side of training images is resized to 800 pixels, and the long side is never more than 1333 pixels. 
% We adopt the multi-scale training for the $3 \times$ schedule, which randomly resizes the short side of the input within the range of [480, 800] while keeping the longer side at most 1333. 
% When testing, the image size setting is the same as the $1\times$.

\begin{figure}[t]
  \centering
  \includegraphics[width=\linewidth]{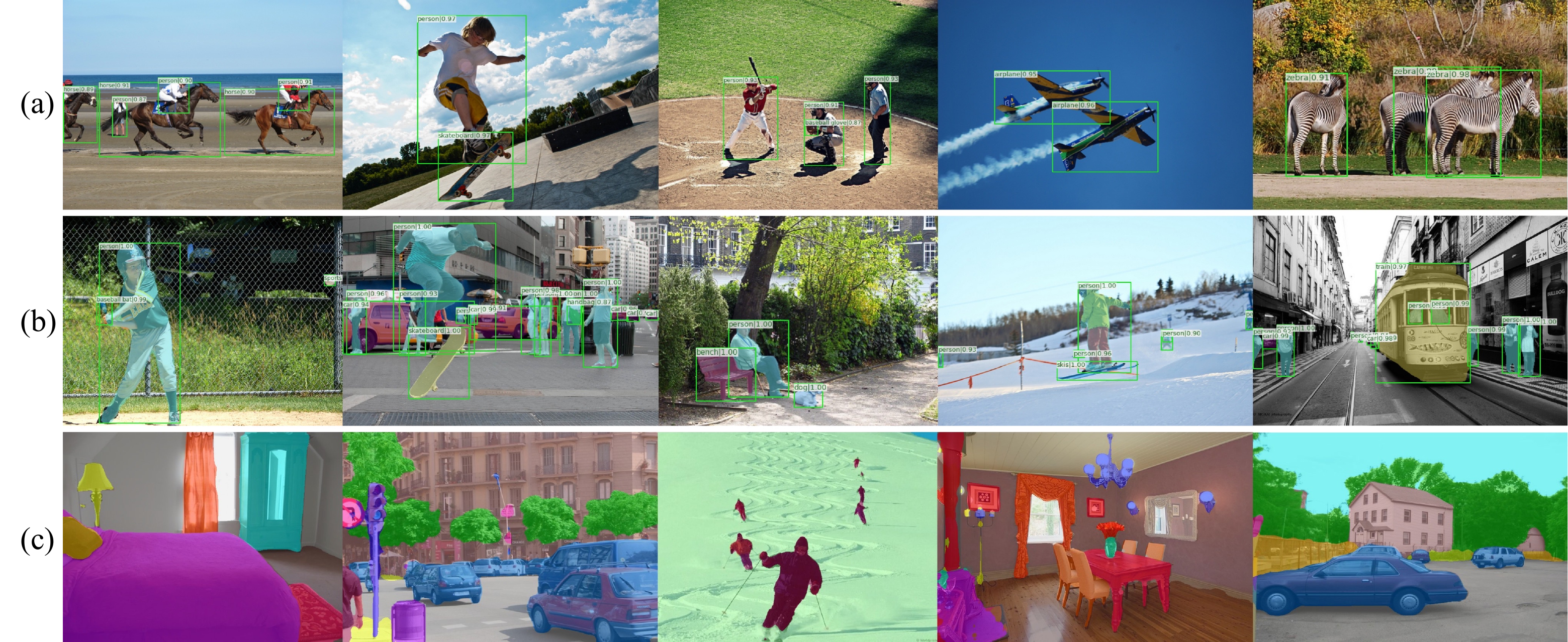}
   \caption{Qualitative results based on ScalableViT-S. (a), (b) and (c) are yielded by  RetinaNet~\cite{RetinaNet}, Mask R-CNN~\cite{MaskRCNN}, and Semantic FPN~\cite{SemanticFPN}, respectively.
   }
   \label{figure:qualitive_results}
\end{figure}

% \vspace*{1\baselineskip}
\noindent \textbf{Result.}
We present results of RetinaNet and Mask R-CNN frameworks in Table~\ref{table:result_COCO}, where $\text{AP}^b$ and $\text{AP}^m$ refer to box mAP and mask mAP, respectively. 
% Additionally, some qualitative results of Mask R-CNN are shown in \cref{fig:QualitativeResult}(a).
For object detection with RetinaNet, ScalableViT performs a notable advantage against its CNN and Transformer counterparts. 
% To be more specific, 
With the $1\times$ schedule, our ScalableViT brings 7.3-8.9 $\text{AP}^b$ against ResNet at comparable settings. Compared with the popular Swin and Twins Transformers, our ScalableViT performs 3.5-3.7 $\text{AP}^b$ and 0.5-2.2 $\text{AP}^b$ improvements, respectively. 
With the $3\times$ schedule, our ScalableViT still achieves competitive performance. 
For Mask R-CNN, our ScalableViT-S outperforms ResNet-50 by 7.8 $\text{AP}^b$ and 7.3 $\text{AP}^m$ with the $1\times$ schedule. ScalableViT-S achieves 3.6 $\text{AP}^b$ and 2.6 $\text{AP}^m$ gains than Swin-T. With the $3\times$ schedule, ScalableViT-S brings 7.7 $\text{AP}^b$ and 6.5 $\text{AP}^m$ against ResNet-50. Similarly, it also surpasses Swin-T and Twins-SVT-S Transformers. Under base level, there is also a similar improvement, demonstrating its stronger context-oriented generalization. 
Additionally, Fig.~\ref{figure:qualitive_results} depicts some qualitative object detection and instance segmentation results from ScalableViT-S-based RetinaNet and Mask R-CNN, which show that contextual representation from the backbone enables the model to detect objects better.
% Under the base outperforms 
% Swin-T by +3.5 $\text{AP}$ under $1\times$ schedule and +3.9 $AP$ under $3\times$ schedule. For the base model, there is also a similar performance boost. Our ScalableViT also achieves competitive performance when it comes to object detection and instance segmentation using Mask R-CNN.

% However, it has a modest disadvantage compared to CSWin, because CSWin does not adopt a spatial reduction manner in self-attention.  Our ScalableViT reduces the spatial dimension of $K$ and $V$ for a better context-oriented generation, which inevitably losses details (like edges).
% This fine-grained information is more critical in pixel-level dense prediction tasks than image-level classification.
% For the dense prediction tasks, the proposed method may be further improved by replacing the reduction manner with the aggregation method~\cite{T2T}. 
% In the future, the reduction manner may be replaced by the aggregation method~\cite{T2T} for better dense predict results.
% Although our model does not achieve astounding results such as classification on the object detection and instance segmentation, it does exhibit some competitiveness.

%-------------- figure visualize featmap
%-------------- figure visualize featmap

%-------------------------------------------------------------------------
\subsection{Semantic Segmentation on ADE20K}
\noindent \textbf{Settings.}
Semantic segmentation experiments are conducted on the challenging ADE20K~\cite{ADE20K} dataset. We use the typical Semantic FPN~\cite{SemanticFPN} and the UperNet~\cite{UperNet} as segmentation frameworks to evaluate our models. 
% Following the common practice, 
We use the MMSegmentation~\cite{mmseg2020} to implement all related experiments, and the settings follow \cite{Twins,swin,PVT}. 
For the Semantic FPN, we train 80K iterations with a batch size 16 on 4 GPUs. For the UperNet, we train 160K iterations with a batch size 16 on 8 GPUs.
% We also use the test time augmentation, including multi-scale test and flip, for better results.
% For the Semantic FPN, we train 80K iterations with a batch size 16 on 4 GPUs. 
% The polynomial policy schedules the learning rate with a power of $0.9$. For the UperNet, we train 160K iterations with a batch size 16 on 8 GPUs. The polynomial policy schedules the learning rate with a power of $1.0$. 
During training, we first resize the short side of input images to 512 pixels, and the long side is never more than 2048 pixels, then they are randomly cropped to $512 \times 512$.
During testing, we resize input images as the training phase but without cropping. We also use the test time augmentation for UperNet, including multi-scale test ($[0.5, 0.75, 1.0, 1.25, 1.5, 1.75] \times$ resolution) and flip.
% More details are in the Appendix~\ref{sec:settings}. 

% \vspace*{1\baselineskip}
\noindent \textbf{Result.}
Table~\ref{table:result_ADE20K} reports the segmentation results.
% obtained using the Semantic FPN and UperNet frameworks. 
For the Semantic FPN, our ScalableViT outperforms Swin Transformer by +3.4 mIoU, +3.2 mIoU, and +3.4 mIoU, respectively, under three FLOPs levels. Compared with CrossFormer-S~\cite{crossformer}, ScalableViT-S performs a modest mIoU but has a fewer computation.
When equipped into the UperNet, the ScalableViT achieves +4 mIoU, +1.9 mIoU, and +1.6 mIoU gains than Swin Transformer under different model sizes. The same competitive results are achieved when test time augmentation is adopted.
In addition, ScalableViT-S outperforms CrossFormer-S by +0.9 mIoU and achieves comparable performance on the base and large size. 
% Although ScalableViT achieves a modest performance under the base and large size compared with CrossFormer, there is still significant competitiveness. 
Fig.~\ref{figure:qualitive_results}(c) shows some qualitative results from ScalableViT-S-based Semantic FPN on validation split. These results indicate that the ScalableViT can obtain high-quality semantic segmentation results under contextual-oriented generalization.
% --------------------------------- table ADE20K
\begin{table}[t]
\centering
\caption{Results on ADE20K segmentation using the Semantic FPN~\cite{SemanticFPN} and UperNet~\cite{UperNet} framework. FLOPs are measured at resolution $512\times2048$. MS refers to the test time augmentation, including flip and multi-scale test.}
\label{table:result_ADE20K}
\resizebox{0.9\textwidth}{!}{
\begin{tabular}{@{}l|ccc|ccc@{}}
\toprule
\multirow{2}{*}{Backbone} & \multicolumn{3}{c|}{Semantic FPN 80k} & \multicolumn{3}{c}{UperNet 160k}           \\
                          & \#Param. & FLOPs & mIoU(\%)      & \#Param. & FLOPs & mIoU/MS mIoU(\%)   \\ \midrule
ResNet50~\cite{ResNet}                  & 29M       & 183G      & 36.7          & ---        & ---      & ---/---            \\
PVT-Small~\cite{PVT}                 & 28M       & 161G      & 39.8          & ---        & ---      & ---/---            \\
Swin-T~\cite{swin}                    & 32M       & 182G      & 41.5          & 60M       & 945G      & 44.5/45.8          \\
Twins-SVT-S~\cite{Twins}               & 28M       & 144G      & 43.2          & 54M       & 901G      & 46.2/47.1          \\
CrossFormer-S~\cite{crossformer}                   & 34M       & 221G      & \textbf{46.0} & 62M       & 980G      & 47.6/48.4 \\
\textbf{ScalableViT-S(ours)}       & 30M          & 174G        & 44.9          & 57M          & 931G        & \textbf{48.5}/\textbf{49.4}          \\ \midrule
ResNet101~\cite{ResNet}                 & 48M       & 260G      & 38.8          & 86M       & 1092G     & ---/44.9           \\
PVT-Medium~\cite{PVT}                & 48M       & 219G      & 41.6          & ---        & ---      & ---/---            \\
Swin-S~\cite{swin}                    & 53M       & 274G      & 45.2          & 81M       & 1038G     & 47.6/49.5          \\
Twins-SVT-B~\cite{Twins}               & 60M       & 261G      & 45.3          & 89M       & 1020G     & 47.7/48.9          \\
CrossFormer-B~\cite{crossformer}                   & 56M       & 331G      &47.7 & 84M       & 1090G     & \textbf{49.7}/\textbf{50.6} \\
\textbf{ScalableViT-B(ours)}       & 79M          & 270G        & \textbf{48.4}         & 107M          & 1029G        & 49.5/50.4          \\ \midrule
ResNeXt101-64$\times$4d~\cite{ResNeXt}             & 86M       & ---        & 40.2          & ---          & ---        & ---/---                  \\
PVT-Large~\cite{PVT}                    & 65M       & 283G      & 42.1          & ---          & ---        & ---/---                  \\
Swin-B~\cite{swin}                       & 91M       & 422G      & 46.0          & 121M        & 1188G     & 48.1/49.7          \\
Twins-SVT-L~\cite{Twins}                  & 104M      & 404G      & 46.7          & 133M        & 1164G     & 48.8/50.2          \\
CrossFormer-L~\cite{crossformer}                      & 95M       & 497G      & 48.7 & 126M      & 1258G     & \textbf{50.4}/\textbf{51.4} \\
\textbf{ScalableViT-L(ours)} & 105M          & 402G        & \textbf{49.4}          & 135M          & 1162G        & 49.8/50.7          \\ \bottomrule
\end{tabular}}
\end{table}
% --------------------------------- table ADE20K

\subsection{Ablation Study}
\label{sec:ablation_study}
% To understand our ScalableViT better, we ablate each critical design by evaluating their performance on ImageNet-1K classification. 
% All hyperparameters required for training are consistent with the ScalableViT-S.

% \vspace*{1\baselineskip}
\noindent \textbf{Analysis for Self-Attention mechanisms.}
Our ScalableViT contains two important designs: SSA and IWSA. We ablate their benefits in Table~\ref{table:result_ablation_1}.
Firstly, all attention modules in ScalableViT-S are replaced with the regular window-based self-attention (WSA).
% , which equals to the $7\times7$ depth-wise convolution with dynamic weights. 
Although WSA achieves $82.4\%$ top-1 accuracy, the dispersed feature (see Fig.~\ref{figure:featmap}) hinders it from better performance on the downstream visual task.
Then, we substituted all attention modules with our IWSA and SSA, respectively. Both of them outperform WSA $0.4\%$ top-1 accuracy. More importantly, they bring $+4.6$ mIoU and $+5.5$ mIoU improvements on ADE20K because of the ability modeling long-range dependency. With spatial scalability ($r_c \equiv 1$), SSA only achieve $82.6\%$ top-1 accuracy and $43.7$ mIoU. Thus, the context-oriented generalization from the cooperation between spatial and channel scalability plays a critical role in visual tasks.
Additionally, we examine the topology by rearranging IWSA and SSA. Results demonstrate that prioritizing IWSA followed by SSA performs best. We also compare IWSA with SWSA~\cite{swin} in ScalableViT, where our IWSA is more appropriate than SWSA.

\noindent \textbf{Speed analysis.} Following~\cite{swin}, we measure throughput of the ScalableViT-S on single 3090 GPU with a batch size of 64 in Table~\ref{table:result_speed}. ScalableViT-S achieves $859.0$ $img/s$, which perform better speed-accuracy trade-offs than Swin-S.

% \vspace*{1\baselineskip}
\noindent \textbf{Effectiveness of Local Interactive Module.}
We examine the effectiveness of LIM in Table~\ref{table:result_ablation_2}.
The ScalableViT-S without position encoding generator (PEG), locally enhanced module (LEM), or LIM is regarded as a baseline model which achieves $82.7\%$ top-1 accuracy on ImageNet.
Then, three modules are inserted and yield $+0.2\%$, $+0.1\%$, and $+0.3\%$ gains than baseline, respectively.
It demonstrates that the reasonable convolution can help the model perform better.
Due to the window connection, LIM outperforms LEM by $+0.2\%$ top-1 accuracy, proving the significance of the information interaction. 
Additionally, we combine PEG with LEM or LIM, whose results are better than only using a single module. Note that the combination of PEG and LIM outperforms the PEG and LEM under the same overhead.
% , confirming the usefulness and effectiveness of our LIM again.
LIM aims to bring global perception into the single Transformer block. Its effectiveness is greatly demonstrated on downstream tasks. Using Semantic FPN with ScalableViT-S on ADE20K, LIM obtains $+2.0$ mIoU, and associating PEG with LIM brings $+3.2$ mIoU gains.

% \vspace*{1\baselineskip}

\begin{table}[t]
\centering
\caption{Ablation study for different self-attention mechanisms and LIM using ScalableViT-S. Top-1 refers to top-1 accuracy (\%) on ImageNet-1K. Semantic segmentation results are yielded from Semantic FPN on ADE20K.}
\label{tab:Ablation}
% ------- table 1
\begin{subtable}{0.45\textwidth}
\centering
\caption{Analysis for different self-attention mechanisms.}
\label{table:result_ablation_1}
\resizebox{\textwidth}{!}{
\begin{tabular}{@{}c|cc|cc@{}}
\toprule
Method       & \#Param. & FLOPs & Top-1 &mIoU(\%)\\ \midrule
WSA         & 30M        & 4.3G      & 82.4      &38.9 \\
IWSA        & 30M        & 4.3G      & 82.9       &43.5\\
SSA          & 34M        & 4.1G      & 82.9       &44.4\\
SSA ($r_c\equiv1$)          & 32M        & 3.9G      &82.6 &43.7      \\
SWSA \& SSA & 32M        & 4.2G      & 82.9      & --- \\
SSA \& IWSA & 32M        & 4.2G      & 83.0      & --- \\ 
IWSA \& SSA & 32M        & 4.2G      & \textbf{83.1}       &\textbf{44.9}\\
\bottomrule
\end{tabular}
}
\end{subtable}
% ------- table 2
\begin{subtable}{0.52\textwidth}
\centering
\caption{Analysis for local interactive module and positional encoding generator.}
\label{table:result_ablation_2}
\resizebox{\textwidth}{!}{
\begin{tabular}{@{}ccc|cc|cc@{}}
\toprule
\multicolumn{1}{l}{PEG} & \multicolumn{1}{l}{LEM} & \multicolumn{1}{l}{LIM} & \multicolumn{1}{|l}{\#Param.} & \multicolumn{1}{l|}{FLOPs} & \multicolumn{1}{l}{Top-1} & \multicolumn{1}{l}{mIoU(\%)} \\ \midrule
                &                        &                        & 32M            & 4.2G         & 82.7      & 41.7                          \\
\checkmark      &                        &                        & 32M            & 4.2G         & 82.9      & 43.2                       \\
                & \checkmark             &                        & 32M            & 4.2G         & 82.8      & ---               \\
                &                        & \checkmark             & 32M            & 4.2G         & \textbf{83.0}  & \textbf{43.7}                         \\
\checkmark      & \checkmark             &                        & 32M            & 4.2G         & 83.0      & ---                       \\
\checkmark      &                        & \checkmark             & 32M            & 4.2G        & \textbf{83.1}    & \textbf{44.9}              \\ \bottomrule
\end{tabular}
}
\end{subtable}
\end{table}

\begin{table}[t]
\centering
\caption{Comparison speed with state-of-the-art models.}
\label{table:result_speed}
\resizebox{0.7\textwidth}{!}{
\begin{tabular}{@{}c|cc|c|c@{}}
\toprule
Method       & \#Param. & FLOPs & Top-1 &throughput($img/s$)\\ \midrule
Swin-T~\cite{swin} & 29M        & 4.5G      & 81.3      & 975.0 \\
Swin-S~\cite{swin} & 50M        & 8.7G      & 83.0      & 589.2 \\
CrossFormer-S~\cite{crossformer} & 31M        & 4.9G      & 82.5      & 859.0 \\
ScalableViT-S(\textbf{ours}) & 32M        & 4.2G      & \textbf{83.1}       &832.9\\
\bottomrule
\end{tabular}
}
\end{table}

\section{Conclusion}
\label{sec:conclusion}
% In this paper, we have presented a vision transformer backbone named ScalableViT which composed of two highly effective self-attention mechanisms (SSA and IWSA). SSA employs two cooperated scaling factors in spatial and channel dimensions for context-oriented generalization, which maintains more contextual cues and learns graphic representations.
% % and pushes the whole framework into a more effective trade-off state between precision and cost.
% IWSA develops a local interactive module to establish information connections between independent windows. Both of them owns the capability to model long-range dependency in a single layer. 
% The proposed ScalableViT alternately stakes these two self-attention modules. It pushes the whole framework into a more effective trade-off state, and achieves state-of-the-art performance on various vision tasks.
In this paper, we have presented a Vision Transformer backbone named ScalableViT, composed of two highly effective self-attention mechanisms (SSA and IWSA). SSA employs two cooperated scaling factors in spatial and channel dimensions for context-oriented generalization, which maintains more contextual cues and learns graphic representations. IWSA develops a local interactive module to establish information connections between independent windows. Both of them owns the capability to model long-range dependency in a single layer. 
The proposed ScalableViT alternately stakes these two self-attention modules. It pushes the whole framework into a more effective trade-off state and achieves state-of-the-art performance on various vision tasks.

\section*{Acknowledgements}
This work was supported by the National Key R\&D Program of China 505 (Grant No.2020AAA0108303), the National Natural Science Foundation of China (Grant No.41876098) and the Shenzhen Science and Technology Project (Grant No.JCYJ20200109143041798).

% appendix
\appendix
\section{Additional Analyses for IWSA}
\begin{figure}[t]
  \centering
  \includegraphics[width=\linewidth]{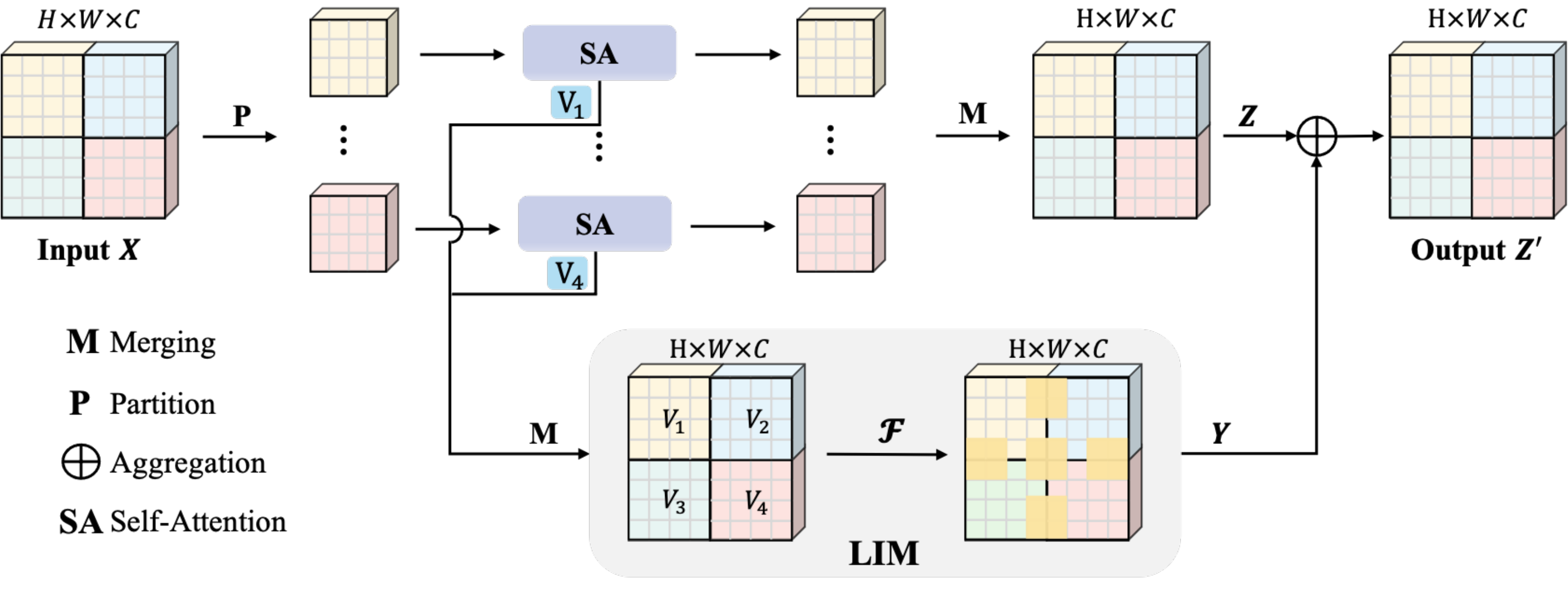}
   \caption{Interactive Window-based Self-Attention (IWSA). Besides the proposed LIM, other parts compose the WSA. The LIM extracts a set of discrete $value$ matrices $\{V_1, V_2, V_3, V_4 \}$ from WSA and merges them via a fusion function $\mathcal{F}$. The output $Y$ is added on $Z$ for an output $Z^{\prime}$ with information interaction.}
   \label{figure:IWSA}
\end{figure}

As shown in Fig.~\ref{figure:IWSA}, IWSA is composed of a window-based self-attention (WSA) and a local interactive module (LIM). WSA splits the global self-attention into many limited windows and yields a collection of discrete $value$ matrices.
LIM build connections between these $value$ matrices through a fusion function $\mathcal{F}$. In practice, this function is replaced with a $3\times3$ depth-wise convolution. 
Additionally, WSA can be viewed as a $7\times7$ depth-wise convolution with an adaptive weight. Thus, $\mathcal{F}$ brings information exchange through a kind of interleaving effect (illustrated by yellow squares in Fig.~\ref{figure:IWSA}). This parallel stagger makes IWSA realize a global receptive field in a single layer.

In Table~\ref{tab:LIM}, we compare the LIM and the LEM on the ADE20K~\cite{ADE20K} using Semantic FPN~\cite{SemanticFPN} framework. All settings are recorded in the Section~\ref{sec:settings}. ScalableViT-S with the LIM achieves $+3.8$ mIoU than the LEM under the same overhead because IWSA can model the long-range dependency in single layer. This result also proves that the global receptive field plays a more critical role on the downstream vision task. Moreover, the LIM can be expanded to other window-based self-attention with different window division styles.

\begin{table}[]
\centering
\caption{LIM vs. LEM on ADE20K using Semantic FPN. \#Param. refers to total parameters of Semantic FPN based on ScalableViT-S backbone. FLOPs are measured at resolution $512\times2048$.}
\label{tab:LIM}
% \resizebox{0.8\textwidth}{!}{
\begin{tabular}{@{}lcccc@{}}
\toprule
Model                & \#Param. & FLOPs & Top-1 & mIoU(\%) \\ \midrule
ScalableViT-S w. LEM & 30M      & 174G  & 83.0  & 41.1     \\
ScalableViT-S w. LIM & 30M      & 174G  & \textbf{83.1}  & \textbf{44.9}     \\ \bottomrule
\end{tabular}
% }
\end{table}

\begin{figure}[t]
  \centering
  \includegraphics[width=0.9\linewidth]{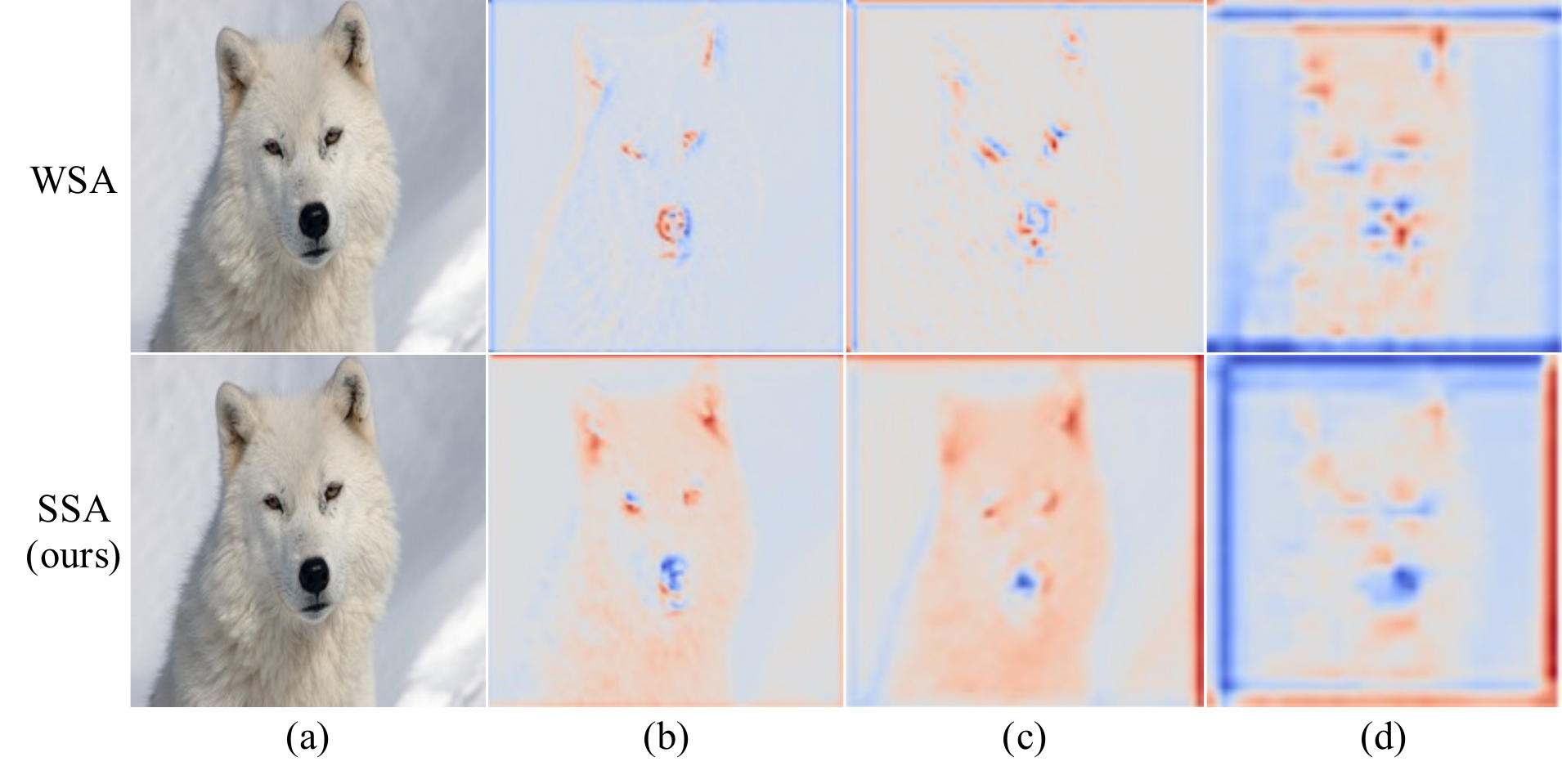}
   \caption{Visualization for feature maps of other blocks. (b), (c), and (d) are output features of the 2nd, 4th, and 24th blocks, respectively.}
   \label{figure:feat_vis_24th}
\end{figure}

\section{Comparing visualizations from other blocks}
We visualize the feature maps after the 2nd, 4th, and 24th blocks in Figure~\ref{figure:feat_vis_24th}. In the 2nd and 4th blocks, the WSA focuses on local regions, especially the ears and nose. 
In the latter 24th block, the WSA attends to contextual information but losses some semantic cues.
Since feature aggregation from a large downsampling ratio ($16$) causes the foreground and background to be poorly separated.
By contrast, the SSA can retain a trail of details although the feature map of the later block are not as continuous as the earlier ones.

\section{More Implementary Details}
\label{sec:settings}
% \vspace*{1\baselineskip}
\noindent \textbf{Classification.}
The classification settings mainly follow DeiT~\cite{Deit}. All variants are trained under a resolution of $224\times224$.
During training from scratch, we employ the AdamW optimizer~\cite{AdamW} with a weight decay of 0.05 and a momentum of 0.9 to train models for 300 epochs. The learning rate is set to 0.001 initially and varies with the cosine scheduler, where a 5-epochs linear warm-up is used to stabilize training. The global batchsize is set to 1024 on 8 V100 GPUs.
Moreover, we apply data augmentations and regularizations, including random cropping, random horizontal flipping~\cite{RandomHorizontalFilpping}, mixup~\cite{Mixup}, CutMix~\cite{CutMix}, random erasing~\cite{RandomEarsing}, label-smoothing~\cite{LabelSmoothing}, stochastic depth~\cite{StochasticDepth}, and repeated augmentation~\cite{RepeatedAugmentaiton}. 
For stochastic depth augmentation, we set the drop rate to $0.2$, $0.5$, and $0.5$ for ScalableViT-S, ScalableViT-B, and ScalableViT-L, respectively.
During testing on the validation set, the shorter side of an input image is first resized to 256, and a center crop of 224 × 224 is used to evaluate the classification accuracy.

\vspace*{1\baselineskip}
\noindent \textbf{Object Detection.} 
We adopt RetinaNet~\cite{RetinaNet} and Mask R-CNN~\cite{MaskRCNN} detection frameworks on COCO~\cite{COCO} that contains 118K training images and 5K validation images. Before training, we initialize the backbone with the weight pre-trained on ImageNet-1K, FPN with Xavier~\cite{Xavier} scheme, and other new layers with Normal scheme ($std=0.01$).
All models utilize AdamW~\cite{AdamW} optimizer, 500-iteration warm-up, $1\times$ (12 epochs), and $3\times$ (36 epochs) schedule with a global batch size of 16 on 8 GPUs. Settings of initial learning rate and weight decay are shown in Table~\ref{tab: lr setting}. For $1 \times$ schedule, the short side of training images is resized to 800 pixels, and the long side is never more than 1333 pixels. The learning rate is declined at the $8th$ and $11th$ epoch with a decay rate of 0.1. For the $3\times$ schedule, we adopt the multi-scale training, which randomly resizes the short side of the input images within the range of [480, 800] while keeping the longer side at most 1333. The learning rate is declined at the $27th$ and $33rd$ with a decay rate of 0.1. When testing, the image size is set as the same as the $1\times$ schedule.

\vspace*{1\baselineskip}
\noindent \textbf{Semantic Segmentation.}
Semantic segmentation experiments are conducted on the challenging ADE20K~\cite{ADE20K}, with 20K images for training and 2K images for validation. We use the typical Semantic FPN~\cite{SemanticFPN} and UperNet~\cite{UperNet} as segmentation frameworks to evaluate our models. Following the common practice, we use the MMSegmentation~\cite{mmseg2020} to implement all related experiments.
We employ the AdamW~\cite{AdamW} to optimize two models. The initial learning rate and weight decay are shown in Table~\ref{tab: lr setting}.
For the Semantic FPN, we train 80K iterations with a batch size 16 on 4 GPUs. The polynomial policy schedules the learning rate with a power of $0.9$. For the UperNet, we train 160K iterations with a batch size 16 on 8 GPUs. The polynomial policy schedules the learning rate with a power of $1.0$. 
During training, we first resize the short side of input images to 512 pixels, and the long side is never more than 2048 pixels, then randomly crop to $512 \times 512$.
During testing, we resize input images the same as the training phase but without cropping. We also use the test time augmentation for UperNet, including multi-scale test ($[0.5, 0.75, 1.0, 1.25, 1.5, 1.75] \times$ resolution) and flip, for better results.

\begin{table}[t]
\centering
\caption{Settings of the initial learning rate and weight decay.}
\label{tab: lr setting}
\resizebox{0.7\textwidth}{!}{
\begin{tabular}{@{}lccc@{}}
\toprule
\multicolumn{1}{l|}{Model}          &\multicolumn{1}{c|}{\#lr scheduler} & \multicolumn{1}{c|}{learning rate} & weight decay \\ \midrule
\multicolumn{4}{c}{Object Detection}                                         \\ \midrule
\multicolumn{1}{l|}{RetinaNet($1 \times$)} &\multicolumn{1}{c|}{Multi-step}  & \multicolumn{1}{c|}{$1 \times 10^{-4}$}  & $1 \times 10^{-4}$         \\
\multicolumn{1}{l|}{RetinaNet($3 \times$)} &\multicolumn{1}{c|}{Multi-step}  & \multicolumn{1}{c|}{$1 \times 10^{-4}$}  & $5 \times 10^{-2}$       \\
\multicolumn{1}{l|}{Mask R-CNN($1 \times$)} &\multicolumn{1}{c|}{Multi-step} & \multicolumn{1}{c|}{$2 \times 10^{-4}$}  & $1 \times 10^{-4}$       \\
\multicolumn{1}{l|}{Mask R-CNN($3 \times$)} &\multicolumn{1}{c|}{Multi-step} & \multicolumn{1}{c|}{$1 \times 10^{-4}$}  & $5 \times 10^{-2}$         \\ \midrule
\multicolumn{4}{c}{Semantic Segmentation}                                    \\ \midrule
\multicolumn{1}{l|}{Semantic FPN}   &\multicolumn{1}{c|}{Polynomial} & \multicolumn{1}{c|}{$1 \times 10^{-4}$}  & $1 \times 10^{-4}$       \\
\multicolumn{1}{l|}{UperNet}        &\multicolumn{1}{c|}{Polynomial} & \multicolumn{1}{c|}{$6 \times 10^{-5}$}  & $1 \times 10^{-2}$         \\ \bottomrule
\end{tabular}
}
\end{table}

\clearpage
\bibliographystyle{splncs04}
\bibliography{main}
\end{document}